\documentclass[11pt]{article}
\usepackage[utf8]{inputenc}

\title{IRT}
\author{ziheng.chen }
\date{October 2019}

\usepackage{amsmath}
\usepackage{graphicx}
\usepackage{geometry}
\usepackage{authblk}
\title{Item Response Theory based Ensemble in Machine Learning}
\author{Ziheng Chen and Hongshik Ahn}
\affil{Department of Applied Mathematics and Statistics, Stony Brook University}
\begin{document}
\setlength{\baselineskip}{25pt}
\maketitle

\begin{abstract}
In this article, we propose a novel probabilistic framework to improve the accuracy of a weighted majority voting algorithm. In order to assign higher weights to the classifiers which can correctly classify hard-to-classify instances, we introduce the Item Response Theory (IRT) framework to evaluate the samples' difficulty and classifiers' ability simultaneously. Three models are created with different assumptions suitable for different cases. When making an inference, we keep a balance between the accuracy and complexity. In our experiment, all the base models are constructed by single trees via bootstrap. To explain the models, we illustrate how the IRT ensemble model constructs the classifying boundary. We also compare their performance with other widely used methods and show that our model performs well on 19 datasets. 
\end{abstract}
\section{Introduction}

Classification ensembles are increasingly gaining attention from the area of machine learning,
especially when we focus on improving the accuracy. The most important feature distinguishing
the ensemble learning from other types of learning is that it combines the predictions from
a group of classifiers rather than depending on a single classifier\cite{zhou2015ensemble}. It is proved in many cases
that the aggregated performance metrics, such as bagging, boosting and incremental learning
outperform others without a collective decision strategy.

If one had to identify an idea as central and novel to ensemble learning, it is the combination rule,
which can be characterized in two ways: simple majority voting and weighted majority voting.
Simple majority voting is just a decision rule which combines the decisions of the classifiers in the
ensemble\cite{zhou2015ensemble}. It is widely applied in ensemble learning due to its simplicity and applicability\cite{lam1997application}.
Weighted majority voting can be done by multiplying a weight to the decision of each classifier
to reflect its ability, and then make the final decision by combining the weighted decisions\cite{rahman2002multiple}.
These two methods utilize the ability of classifiers based on their performance on training the data.
Thus it does not require any parameter tuning once the individual classifiers have been trained.

Here we propose a novel probabilistic framework for the weighted voting classification ensemble.
We treat each data point as a problem and different classifier as a subject taking an exam in class.
As we know, the performance of one student on a problem depends on two major factors: difficulty of
the problem and competence of the student\cite{kim2011weight}. In the training data,
some have significant features
and are easy to classify, whereas some are hard-to-classify cases because they are near class
boundaries. Thus, similar to an exam in class, we define the competence of a classifier as
the capability of correctly classifying difficult cases, rather than the number of correctly
classified cases. For instance, suppose a classifier correctly classifies some easy cases but fails
to deal with difficult cases. Another classifier correctly classifies some difficult cases,
while incorrectly classifies easy cases. Then it makes sense that a higher weight is given to
the second classifier than the first one.

In this paper, we propose a method which can simultaneously evaluate the ability of a classifier
and difficulty of classifying a case. Here, we employ the IRT (Item Response Theory) framework\cite{embretson2013item},
which is widely applied to
psychological and educational research, to estimate the latent ability of classifiers.

\section{Motivation and Background}

\subsection{Classifier's ability }

Classifier $i$'s ability is defined by the parameter $\theta_i$, which measures its capacity to handle different samples. Not only the number of cases it can classify, the hardness of case is also considered when estimating the parameter. The classifier's ability is directly connected with the weight assigned to each classifier in the ensemble, which is a real value between 0 and 1. A classifier having very negative ability leads to a weight close to 0, while the opposite is true for the classifier with very positive ability. Outliers, observations near the boundary and observations surrounded by multiple observations from other class can usually only be correctly classified by classifier with high value of ability.

\subsection{Item Response Theory}

Item response theory considers a set of models that relate responses given to items to latent abilities of the respondents\cite{embretson2013item}. It is wildely applied in educational testing and psychological evaluation. In this model, the probability of a response is a function of the classifier's ability and observation's difficulty. As in our case, one classifier either correctly classify or wrongly classify an observation, we only focus on the dichotomous models. In the original model,their relationship can be described below:

\[P(y_{ij}=1|\theta_i,\beta_j)=\Phi(\theta_i-\beta_j)=\int_{-\infty}^{\theta_i-\beta_j}\frac{1}{\sqrt{2\pi}}e^{\frac{-t^2}{2}}\,dt\]

$y_{ij}$ is a binary response of a classifier i to observation j, where $y_{ij}=1$ for a correct classification and $y_{ij}=0$ otherwise. $\theta_i$ is latent ability parameter for classifier i and $\beta_j$ is the difficulty of observation j. As it only contains one item parameter$\beta_j$, it is named 1PNO. For the 2PNO model and 3PNO model, we will introduce 2 item parameters and 3 item parameters correspondingly. In our first method, we will use the 3PNO as the basic framework. In our second method, we design a 2PNO.

\subsection{Related Works}

A comprehensive framework of majority vote, weighted majority vote, recall combiner and the naive bayes combiner is presented in \cite{kuncheva2014weighted}. Based on their probabilistic framework, four ensemble methods are derived subsequently by progressively relaxing the assumption. They show that four different methods can be generalized in a bayesian decision procedure. They construct the model directly by considering the misclassification probability, rather than the hardness of the observations and the ability of the classifier. In \cite{kabir2018mixed}, although instance hardness is taken into consideration, they didn't discuss the detailed decision mechanism between the classifiers and the observations.

We propose the IRT ensemble to evaluate the hardness of samples and the ability of classifiers simultaneously. This work is similar to the WAVE \cite{kim2011weight} proposed in 2011. However, although they use a similar idea, they didn't statistically explain the weight. Our proposed methods model the weight assignment in a probabilistic way and can explain the corresponding relationship between the classifier's ability and observations' hardness. 

Many prior works focus on assigning the weights to different classifiers constructed by bootstrapping\cite{breiman1996bagging}\cite{penrose1946elementary}. In \cite{winham2013weighted}\cite{chen2017high}, they apply the weighted ensemble into high dimensional cases. Ensemble pruning algorithms are also considered in \cite{zhou2014effective}\cite{zhang2006ensemble}, focusing on pruning the ensembles with significance features. To the best of our knowledge, our method is the first approach that introducing the item response theory into the ensemble learning and gives a statistical explanation of the samples' hardness and classifiers' ability. 

\section{Model Development}

In this section, we introduce the proposed method in detail. We treat the classifiers as competitors
and want to evaluate their performance by considering the accuracy in classifying hard-to-classify samples.
The basic rule is to assign a higher weight to the classifiers with higher accuracy in classification.
Thus, we adopt the framework in the item response theory\cite{martinez2016making}, which is widely applied in the educational
testing to evaluate the items and people simultaneously\cite{zanon2016application}\cite{barbera2015birds}\cite{fu2019intelligent}. We firstly describe our model and explain why it works. Then we introduce two methods to make an inference.

\subsection{Model description}

We consider a set of classifiers $\Omega=\{C_1,C_2, \cdots, C_n\}$ and a set of data points
$\Phi=\{S_1,S_2, \cdots, S_m\}$. For each classifier  $C_i$, there is a corresponding parameter $\theta_i$
to denote the ability of the classifier. Similarly, we assign a difficulty parameter $\beta_j$ for each
data point $S_j$. Based on the IRT framework, the probability of a response for a data point is
a function of the classifier's ability and difficulty of classifying the case. Although there have been
various models developed within the IRT framework, we focus on the basic unidimensional IRT model because
it models the classifier-sample interaction by two single unified traits $\theta$ and $\beta$.
In this model, the response generated from the interaction has only two choices: success or failure.
In our problem, success means the classifier recognizing the label of a sample correctly.

Now we can formulate the model. Suppose we have $k$ classifiers and $n$ data points in a training set.
We denote the $n \times k$ matrix {\bfseries Y} as the performance matrix. For each element $Y_{ij}$,
1 is assigned if classifier $i$ correctly classifies sample $j$, or 0 is assigned otherwise. We also
define {\bfseries l} as an $n \times 1$ vector with $1$'s. Finally, {\bfseries I} is a $k \times k$
identity matrix. For an easy description, we first propose the following two assumptions:

\begin{enumerate}
\item The performance parameter of a classifier and difficulty parameter of a sample are rational numbers,
and the probability of a correct classification can be expressed as a cumulative distribution
function (CDF).
\item The classifiers give their decisions independently conditioned upon the training data.
\end{enumerate}

\noindent
For each classifier $C_i$, the probability of a correct classification for sample point $S_j$
equals:
\noindent
$P(Y_{ij}=1)=\phi(\alpha_j\theta_i-\beta_j)+\gamma_j(1-\phi(\alpha_j\theta_i-\beta_j))$,
\noindent
Similarly, the probability of a wrong classification equals:
\noindent
$P(Y_{ij}=1)=(1-\gamma_j)(1-\phi(\alpha_j\theta_i-\beta_j))$, where $\phi(x)$ is a CDF. Now we explain extra parameters and their original purpose.

\begin{enumerate}
\item The discrimination parameter $\alpha_j$ of case $S_j$ reflects the steepness of the probability
function. If we set $\gamma_j=0$ and differentiate the function $\phi$ of $\theta_i$, then
the derivative is $\alpha_j\phi^{'}(\alpha_j\theta_i-\beta_j)$, and $\alpha_j$ serves as the
multiplier. The larger the value of $\alpha_j$, the steeper the probability.
Hence at some point with a large $\alpha_j$, any small improvement of the classifier's ability can make a huge difference in
the response, which can be used to detect subtle differences in the ability of the classifier.
\item We also define $\gamma_j$ as a guessing parameter. There is a small probability that a classifier
can correctly classify the situation without really learning the features from the training data.
\end{enumerate}

We can see the advantage of the model. The performance of a classifier is estimated based on the
responses to discriminating items with different levels of difficulty, but not by the accuracy.
Classifiers which correctly classify the difficult cases will get a high estimated
value of the performance parameter. Hard-to-classify data points tend to be correctly classified
by highly performing classifiers.

\subsection{Inference}

According to the second assumption, we can write the likelihood function as
\[ f(y|\alpha,\gamma)=\prod_{i=1}^n\prod_{j=1}^kP(Y_{ij}=1)^{Y_{ij}}(1-P(Y_{ij}=1))^{1-Y_{ij}}. \]
If we directly optimize the likelihood function, it will be very unstable due to the non-convexity of
the function. Thus, we consider using Markov Chain Monte Carlo (MCMC) to estimate the parameters\cite{gilks1995markov}.
The basic idea of MCMC is to use a series of Markov chains and transition kernel to estimate the
parameters. After the procedure, we obtain an estimation of the parameters. To make a concise notation,
we denote a set of all the parameters as $\Theta$, which includes all the discrimination parameters $\alpha_j$,
the guessing parameters $\gamma_j$, difficulty parameters $\beta_j$ and performance parameters $\theta_i$.
For the training data, we denote the sample space as $\Pi$. Thus the joint probability function can be
written as $f(\Pi|\Theta)$.

\subsubsection{Metropolis Hastings algorithm}

The Metropolis Hastings (M-H) algorithm is a very basic method in the MCMC family, which constructs the
Markov chain from the parameter's posterior distribution with a proposal kernel\cite{chib1995understanding}\cite{junker2016markov}. If we supply
a prior distribution $f(\Theta)$ to the parameters, we can write the joint distribution of
the parameters and data as $f(\Pi,\Theta)=f(\Pi|\Theta) f(\Theta)$, and following the Bayes rule,
the posterior distribution of $\Theta$ as
\[ p(\Theta|\Pi)=\frac{f(\Pi|\Theta)f(\Theta)}{\int f(\Pi|\Theta)f(\Theta)d\Theta}\propto
f(\Pi|\Theta)f(\Theta). \]
Thus, we get the posterior distribution of certain parameter $\Theta$.

Then we grow the Markov chain by sampling from the posterior distribution in multiple steps\cite{kim2007estimating}.
In particular step $i$ of the algorithm, in which the target parameter is $\Theta^i$,
we draw a sample $\Theta^{'}$ from the proposal kernel $q(\Theta^{'}|\Theta^{i})$. The probability of
accepting it as the value of $\Theta^{i+1}$ in next iteration is as follows:
\[ \alpha=\min\left\{\frac{f(\Theta^{'}|\mbox{rest})q(\Theta^{'}|\Theta^{i})}{f(\Theta^{i}|\mbox{rest})
  q(\Theta^{i}|\Theta^{'})},1\right\} \]
Here, the rest of the parameters are denoted as rest. Suppose $u\leq \alpha$. Then we can generate another random number $u$ from uniform distribution U(0,1).
We update $\Theta^{i+1}$ as $\Theta^{'}$. Otherwise, we reject the proposal value.

The above is a brief summary of the M-H algorithm. In practice, we have to estimate
an array of parameters rather than a single parameter, so we need to decompose the parameter vector
into different components and update them one by one through the M-H algorithm. This slows down the M-H algorithm. Besides, the prior
distribution of the parameters should be carefully determined by considering the effectiveness of
the rejection process. Last but not least, the initial value also plays a fundamental role
in the efficiency of estimating the parameters.

In this model, we assign the parameters with the following priors:
\begin{eqnarray*}
\theta_i &\sim&  \mbox{N}(0,\sigma_{i}^{2}) \\
\log(\alpha_j ) &\sim& \mbox{N}(\mu_a,\sigma_a^2) ~ \mbox{(we constrain} ~ \alpha_j \geq 0) \\
\beta_j &\sim&  \mbox{N}(0,\sigma^{2}) \\
\sigma_{i}^{2} &\sim& \mbox{IG}(\alpha_{\theta},\beta_{\theta})
\end{eqnarray*}
Then we derive the posterior probability density function based on the given prior.  We can use Pystan to construct the model and assign priors to the parameters. Pystan also provides us different algorithms to make an inference.
It is obvious that we cannot obtain a concise form of the posterior distribution because the posterior
form does not belong to any exponential family except the last step. Thus, we resort to the M-H algorithm
for the updating of parameters $\theta_i$, $\alpha_j$, $\beta_j$. Finally, we use the updated inverse
gamma distribution to update $\sigma_i$. Details of the parameter setting will be shown in the
following section.

\subsubsection{Gibbs sampling algorithm}

Throughout the paper, vectors or matrices are denoted using boldface.
\[P(Y_{ij}=1)=\phi(\alpha_j\theta_i-\beta_j)+\gamma_j(1-\phi(\alpha_j\theta_i-\beta_j))\]

In order to make inference, we use the data augmentation method \cite{tanner1987calculation}in the likelihood function so as to make it easier to analyze. In many cases when the likelihood function cannot be closely approximated by the normal likelihood, the data augmentation method can simplify it by introducing a series of latent variables\cite{albert1992bayesian}\cite{sheng2008markov}.

\noindent
{\bf Introducing latent variables} \\
In our problem, we define two $n$ by $m$ matrix variables $\textbf{W,\;Z}$ which are associated with 
the generating process of the performance matrix. Before explaining them, we make one assumption:

\noindent
{\bf (Assumption 1)}:
If a sample point $j$  falls in the stable region, which is constructed by the classifier $i$, and is not close to 
the boundary, the classifier $i$ can correctly classify the sample with probability 1.
\begin{eqnarray*}
W_{ij}&=&\left\{
\begin{array}{lr}
1\qquad  \mbox{if sample} ~ j ~ \mbox{is within classifier} ~ i ~ \mbox{'s stable region} \\
0\qquad \mbox{if sample} ~ j ~ \mbox{is close to classifier} ~ i~ \mbox{'s unstable boundary}\\
\end{array}
\right.
\end{eqnarray*}
 \[Z_{ij} \sim N(\eta_{ij},1)\qquad\qquad\qquad \eta_{ij} = \alpha_j \theta_i-\beta_j \]

\noindent
For the relation between {\bf{W}} and {\bf{Z}}, we have the following definition:
\begin{eqnarray*}
\left\{
\begin{array}{lr}
W_{ij} = 1 & \mbox{if} ~ Z_{ij} \geq 0  \\
W_{ij} = 0 & \mbox{if} ~ Z_{ij} < 0
\end{array}
\right.
\end{eqnarray*}
{\bf Prior for normal parameters} \\
For the classifier's ability $\theta_i$, we assign a normal distribution
$\theta_i \sim N \left( \mu,\sigma^2 \right)$\cite{sheng2015bayesian}. As $\alpha_j,\;\beta_j$ are parameters
to describe the discrimination and difficulty of problem $j$, we can stack
$\alpha_j,\;\beta_j$ up for simplicity and denote the vector as ${\bf \phi_j}$.
In accordance with $\theta_i$, we also assign the new vector of a multivariate
normal distribution prior:
\begin{equation*}
{\bf \Phi}={
\left[ \begin{array}{ccc}
{\bf \phi_1}, & {\bf \phi_2}, & \hdots,  {\bf \phi_m}\\
\end{array}
\right ]}
\end{equation*}
\begin{equation*}
{\bf \Sigma_{\phi}}={
\left[ \begin{array}{ccc}
\sigma_{\alpha}^2  &  \rho\sigma_{\alpha}\sigma_{\beta}\\
\rho\sigma_{\alpha}\sigma_{\beta}  &  \sigma_{\beta}^2\\
\end{array}\right]}
\end{equation*}
\begin{equation*}
{\bf \phi_j}={
\left[ \begin{array}{ccc}
\alpha_j\\
\beta_j\\
\end{array}
\right ]}
\quad \sim \quad
{\bf \mbox{N}}\left({\left[ \begin{array}{ccc}
\mu_{\alpha_j}\\
\mu_{\beta_j}\\
\end{array}\right] }\quad , \quad{\bf \Sigma_{\phi}}\right)
\end{equation*}

\noindent
We also assume the parameter
\begin{equation*}
{\bf M_{j}}={\left[\begin{array}{ccc}
                      \mu_{\alpha_j}\\
                      \mu_{\beta_j}\\
                     \end{array}\right] }
                     \quad\sim\quad
 {\bf \mbox{N}}\left({\left[ \begin{array}{ccc}
\tau_{\alpha_j}\\
\tau_{\beta_j}\\
\end{array}\right] }\quad , \quad{\bf I}\right)
\end{equation*}

\noindent
For the hyperparameters $\tau_{\alpha_j},\;\tau_{\beta_j}$, we assign a value related to the number of people who correctly answer problem $j$.

Finally, for the guessing parameter $\gamma$, we also assign a beta distribution
\[\gamma_j\quad\sim\quad \mbox{Beta}(s,t)\]
Thus the joint posterior distribution of ${\bf(\Theta,\Phi,M,\Gamma,W,Z|y)}$ is
\begin{eqnarray*}
& & P{\bf(\Theta,\Phi,M,\Gamma,W,Z|y)} \\
&=& P(y|{\bf W,\Gamma})P{\bf(W|Z)} P {\bf(Z|\Theta,\Phi,M)} P({\bf\Theta})P({\bf \Phi|M})P({\bf M})P({\bf \Gamma})
\end{eqnarray*}

\noindent
{\bf Parameter inference} \\
According to the Gibbs algorithm, we can generate the posterior samples by sweeping through each variables. In each iteration, we sample from the conditional distribution with the remaining values fixed at the current value.

\noindent
{\bf Inference on latent variables}
\begin{eqnarray*}
P(y_{ij}=1, W_{ij}=1|\sim) &=& P(y_{ij}=1| W_{ij}=1)P( W_{ij}=1)=\Phi(\eta_{ij}) \\
P(y_{ij}=1,W_{ij}=0|\sim) &=& P(y_{ij}=1|W_{ij}=0)P(W_{ij}=0)=\gamma_j(1-\Phi(\eta_{ij})) \\
P(y_{ij}=0,W_{ij}=1|\sim) &=& 0 \\
P(y_{ij}=0,W_{ij}=0|\sim) &=& (1-\gamma_j)(1-\Phi(\eta_{ij}))
\end{eqnarray*}

\noindent
Thus, the conditional distribution of $ W_{ij}$ can be derived as follows:
\begin{eqnarray*}
W_{ij}=\left\{
\begin{array}{cl}
\mbox{Bernoulli}\; \left( \frac{\Phi(\eta_{ij})}{\gamma_j+(1-\gamma_j)\Phi(\eta_{ij})} \right) &
  \mbox{if} ~ y_{ij}=1\\
0 &  \mbox{if} ~ y_{ij}=0
\end{array}
\right.
\end{eqnarray*}

\noindent
Similarly, we can obtain the conditional distribution of {\bf Z} as
\begin{eqnarray*}
Z_{ij}&=&\left\{
\begin{array}{ll}
N(\eta_{ij},1){ I(Z_{ij}\geq 0)} & \mbox{if} ~ W_{ij} = 1\\
N(\eta_{ij},1){ I(Z_{ij}< 0)} & \mbox{if} ~ W_{ij} = 0
\end{array}
\right.
\end{eqnarray*}

\noindent
{\bf Inference on normal variables}

\begin{enumerate}
\item  For $\theta_i$,
\begin{eqnarray*}
P(\theta_i|\sim) &\propto& \prod_{j=1}^m
 \exp\left(-\frac{(z_{ij}-\left(\theta_i\alpha_j-\beta_j\right)^2)}{2}\right)
  \exp\left(-\frac{(\theta_i-\mu)^2}{2\sigma^2}\right) \\
 &\propto& \exp\left(-\left( \frac{\sum_{j=1}^m\alpha_j^2}{2}
 +\frac{1}{2\sigma^2}\right)\theta_i^2
 -\left(\sum_{j=1}^m\alpha_j(z_{ij}+\beta_j)+\frac{\mu}{\sigma^2}\right)\theta_i\right)\\
\end{eqnarray*}

Thus we have
\[ \theta_i \sim N \left(\frac{\sum_{j=1}^m(z_{ij}+\beta_j)\alpha_j
+\frac{\mu}{\sigma^2}}{\frac{1}{\sigma^2}+\sum_{j=1}^{m}\alpha_j^2},
\frac{1}{\frac{1}{\sigma^2}+\sum_{j=1}^m\alpha_j^2}\right) \]

\item For ${\bf \Phi_{ \cdot j}}$

if we denote $\overline{X}$ as follows:
\begin{equation*}
\overline{X} ={
\left[ \begin{array}{ccc}
\theta_1 & -1\\
\theta_2 & -1\\
\theta_3 & -1\\
\theta_4 & -1\\
\theta_5 & -1\\
\theta_6 & -1\\
\vdots & \vdots\\
\theta_n & -1
\end{array}
\right ]}
\end{equation*}
Then we have
\begin{equation*}
\overline {\bf E}={
\left[ \begin{array}{cccc}
\eta_{11} & \eta_{12} & \cdots & \eta_{1m}\\
\eta_{21} & \eta_{22} & \cdots & \eta_{2m}\\
\vdots & \vdots & \ddots &  \vdots \\
\eta_{n1} & \eta_{n2} & \cdots &  \eta_{nm}
\end{array}
\right ]}=\overline{X} \Phi
\end{equation*}

\begin{eqnarray*}
P({\bf \Phi_{ \cdot j}}|\sim) &\propto& \exp\left(-\frac{1}{2}({\bf Z_{ \cdot j}}
 -{\bf \overline{X} \Phi_{\cdot j}} )^{T}\right)({\bf Z_{\cdot j}}-{\bf \overline{X} \Phi_{\cdot j}} ) \\
 & & \times
  \exp \left(-\frac{1}{2}({\bf{\Phi_{\cdot j}-M_j}})^{T} \Sigma_{\phi}^{-1}
  ({\bf{\Phi_{\cdot j}-M_j}})\right) \\
 &\propto& \exp \left(-\frac{1}{2} \left({\bf \Phi_{\cdot j}^T \left(\overline{X}^T
  \overline{X} +\Sigma_{\Phi_{\cdot j}}^{-1} \right) \Phi_{\cdot j}}
  -{\bf 2\Phi_j^{T} \left( \overline{X}^{T}Z_{\cdot j}^{T}+\Sigma_{\phi}^{-1}
  \right) M_j }\right) \right)
\end{eqnarray*}

\noindent
Thus, we have
\[ {\bf P(\Phi_{ \cdot j}|\sim) \sim {N \left( \left(\overline X^{T}\overline{X}
 +\Sigma_{\phi}^{-1} \right)^{-1} \left(\overline{X}^{T} Z_{\cdot j}+\Sigma_{\phi}^{-1}M_j \right),
 \left(\overline{X}^{T}\overline{X} +\Sigma_{\phi}^{-1} \right)^{-1}\right)}} \]

\item For ${\bf M_j}$
\begin{eqnarray*}
P({\bf M_j}|\sim) &\propto& \exp\left(-\frac{1}{2}({\bf \Phi_j-M_j})^{T}
 \Sigma_{\phi}^{-1}({\bf \Phi_j-M_j})\right) \\
 & & \times \exp\left(-\frac{1}{2}({\bf M_j-T_j})^{T}({\bf M_j-T_j})\right) \\
 &\propto& \exp\left(-\frac{1}{2}{\bf M_j^{T}(\Sigma_{\phi}^{-1}+I)M_j}
  -2{\bf M_j^{T}(\Sigma_{\phi}^{-1}\Phi_j+T_j)}\right)
\end{eqnarray*}

Thus, we have
\[ {\bf M_j \sim N \left( \left( \Sigma_{\phi}^{-1}+I \right)^{-1}
 \left( \Sigma_{\phi}^{-1}\Phi_{\cdot j}+T_j \right),
 \left(\Sigma_{\phi}^{-1}+I \right)^{-1}\right)} \]

\item For $\gamma_j$
\begin{eqnarray*}
P(\gamma_j|\sim) &\propto& \prod_{j=1}^m\gamma^{\sum_{i=1}^n{\bf I}(W_{ij}=0,y_{ij}=1)}
 (1-\gamma_j)^{\sum_{i=1}^n{\bf I}(W_{ij}=0,y_{ij}=0)}\gamma_j^{s-1}(1-\gamma_j)^{t-1} \\
 &\propto& \gamma^{\sum_{j=1}^m\sum_{i=1}^n{\bf I}(W_{ij}=0,y_{ij}=1)+s-1}
 (1-\gamma_j)^{\sum_{j=1}^m\sum_{i=1}^n{\bf I}(W_{ij}=0,y_{ij}=0)+t-1}
\end{eqnarray*}
\[ \gamma_j \sim \mbox{Gamma}\left(\sum_{j=1}^m\sum_{i=1}^n{\bf I}(W_{ij}=0,y_{ij}=1)+s,\sum_{j=1}^m\sum_{i=1}^n{\bf I}(W_{ij}=0,y_{ij}=0)+t\right) \]
\end{enumerate}

\vspace{3ex}
\section{Bernoulli-Beta Model}

\subsection{Bernoulli-Beta model construction}

  In the previous model, we quantify the classifier's ability by defining the probability of a correct
   classification, and make an assumption about the parameters for simplicity of calculation in the MCMC
   approach. However, this method is limited in that it confines the distribution of a correct classification,  which cannot be directly measured by the performance matrix. To obtain a closed form, we introduce many normal assumptions for the relationship between latent variables. Besides, Gibbs sampling is also time consuming. Thus, instead of assuming
   a certain distribution for the latent variables, a new model is proposed in which a constraint relaxation is
   applied to the latent parameters. In this model, the successful prediction
   of sample $j$ is also determined by its difficulty and the ability of classifier $i$.
  Now, we consider the generating process of each element  $Y_{ij}$ in performance matrix $Y$.
  For each element  $Y_{ij}$:
\begin{eqnarray*}
Y_{ij} &\sim& \mbox{Bernoulli}(P_{ij}) \\
P_{ij} &\sim& \mbox{Beta}(m_{ij},n_{ij}) 
\end{eqnarray*}

  The Bernoulli-Beta conjugate distribution, we can see the value of $Y_{ij}$ is strongly associated with a probability $P_{ij}$, and we define it as the successful parameter. It is clear that the larger the $P_{ij}$, the higher the probability of $Y_{ij}=1$, meaning that the classifier $i$ correctly classifies sample $j$. Their relationship can be shown below:

\[ P(Y_{ij}=1)=P_{ij} \quad\quad\quad f(P_{ij})=\frac{\Gamma(m_{ij}+n_{ij})}
{\Gamma(m_{ij})\Gamma(n_{ij})}P_{ij}^{m_{ij}-1}(1-P_{ij})^{n_{ij}-1} \]

Thus, we want the successful parameters to be increasing functions of  $\Delta_{ij}=\theta_i-\beta_j$ and keep the parameters $m_{ij}$ and $n_{ij}$ positive. We need to construct a special relationship to link them with $\theta_i$ and $\beta_j$. We can construct a function
\[m_{ij}=\exp\bigg\{\frac{\alpha_j\theta_i-\beta_j}{2}\bigg\}\quad\quad\quad n_{ij}=\exp\bigg\{\frac{-\alpha_j\theta_i+\beta_j}{2}\bigg\}\]

The exponential function ensures that the parameters can take on positive values. With this structure, the success probability can be indirectly affected by the classifier's ability and the sample's difficulty. Most importantly, no assumptions are made for the classifier's ability\cite{noel2007beta}.

We can also view the relationship in a different perspective. If we find the expectation of $P_{ij}$ under the assumption and expand it as a function of parameters $\theta_i$ and $\beta_j$, the following sigmoid function will appear.

\[E(P_{ij})=\frac{m_{ij}}{n_{ij}+m_{ij}}=\frac{\exp\bigg\{\frac{\alpha_j\theta_i-\beta_j}{2}\bigg\}}
 {\exp\bigg\{\frac{\alpha_j\theta_i-\beta_j}{2}\bigg\}+\exp\bigg\{\frac{-\alpha_j\theta_i+\beta_j}{2}\bigg\}}
  =\frac{1}{1+\exp\bigg\{-\alpha_j\theta_i+\beta_j\bigg\}}\]

Obviously, we bridge the latent parameter $P_{ij}$ and $\theta_i$, $\beta_j$ in a more flexible approach
than the previous model. Their relation is no longer defined by a constant equation, but a distribution
with a sigmoid function. Although the relation here has no distributional interpretation, it quantifies
how the distance between $\theta_i$ and $\beta_j$ contributes to the successful probability, and then
generates a prediction result\cite{juncai2015prediction}. Moreover, $\theta_i$ and $\beta_j$ are all free variables without
a strong distributional constraint. Figure 1 displays the expected response function with different difficulties and discriminations.

\begin{figure}[h]
	\centering  
	\includegraphics[width=1.1\linewidth]{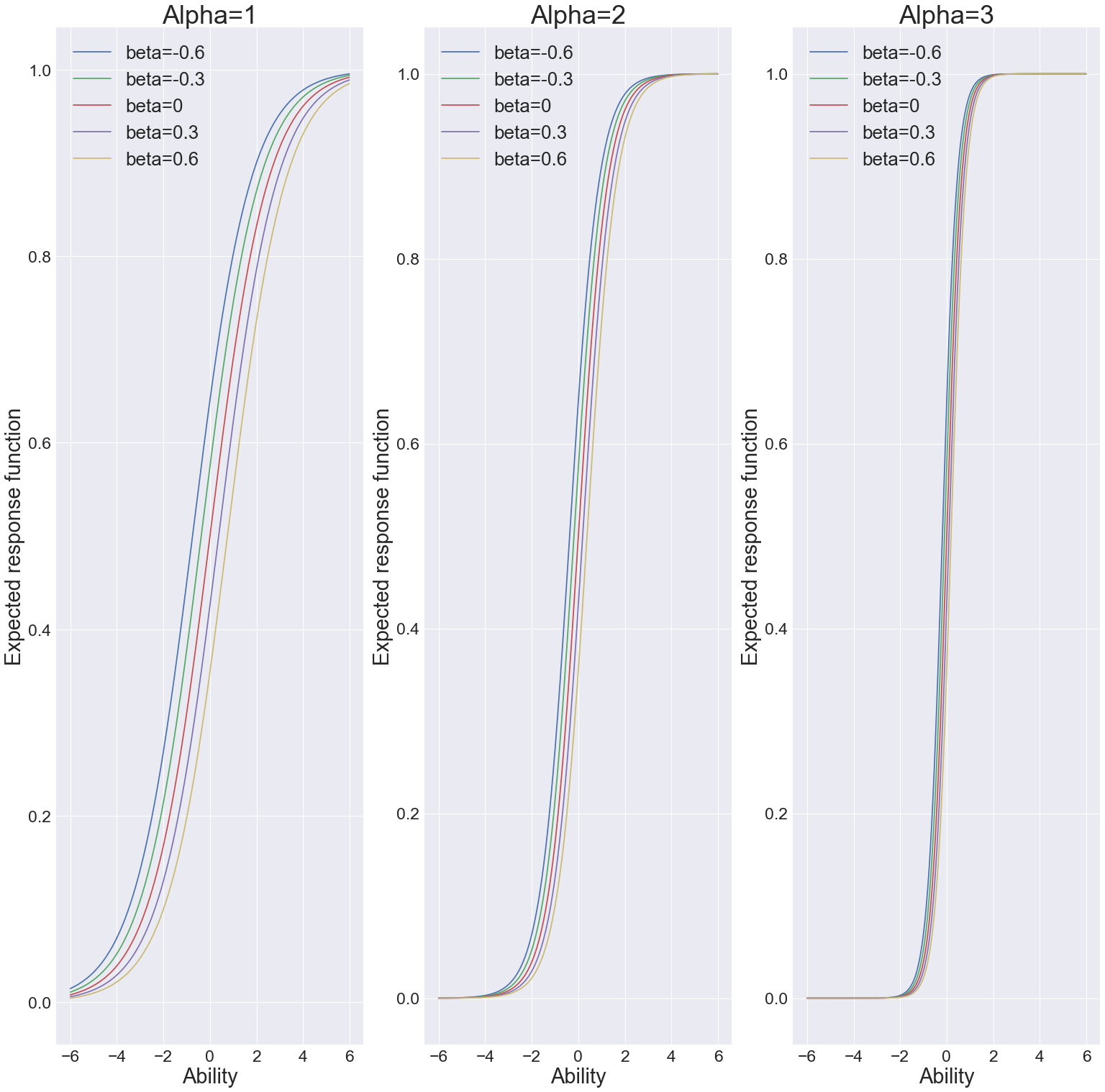}  
	\caption{The expected response function with different difficulty and discrimination. The higher the discrimination, the steeper the curve. The difficulty can affect the probability for correctly classifying the sample}  
	\label{fig:mcmthesis-logo}   
\end{figure}

Now we can summarize the full model joint probability as follows:
\begin{eqnarray*}
P(Y,P,M,N) &=&\prod_{i=1}^n\prod_{j=1}^m
P_{ij}^{y_{ij}}(1-P_{ij})^{1-y_{ij}}\frac{\Gamma(m_{ij}+n_{ij})}{\Gamma(m_{ij})
 \Gamma(n_{ij})}P_{ij}^{m_{ij}-1}(1-P_{ij})^{n_{ij}-1} \\
L(Y,P,M,N) &=&\sum_{i=1}^n\sum_{j=1}^m (y_{ij}+m_{ij}-1) \ln(P_{ij})+(n_{ij}-y_{ij})\ln(1-P_{ij}) \\
  & & +\ln \left( \frac{\Gamma(m_{ij}+n_{ij})}{\Gamma(m_{ij})\Gamma(n_{ij})} \right)\\
&=&\quad \sum_{i=1}^n\sum_{j=1}^m\quad L_{ij}
\end{eqnarray*}
We denote all the parameters $m_{ij}$ and $n_{ij}$ as $\Theta=\bigcup_{i,j} \left(m_{ij},n_{ij}\right)$
 Now the model is constructed.

\subsection{Parameter inference}

  A more difficult, but common, situation is that we introduce a latent parameter $P_{ij}$ while making
  no distributional assumption about parameters $\theta_i$ and $\beta_j$. To solve this problem, we adopt
  the EM algorithm\cite{li2005regularized}\cite{borman2004expectation} which is a standard tool for the maximum likelihood algorithm with latent variables\cite{deng2017novel}.
  We notice that the parameters are redundant because the likelihood function only depends on the
  distance of $\theta_i$ and $\beta_j$. Thus, a constraint for the parameters is necessary. We set the
  mean of $\beta_j$ equal to 0, i.e.,  $\sum_{j=1}^m\beta_j=0$.

\noindent
{\bf E-step} \\
The E-step, on the ($k+1$)th iteration, requires the calculation of
\begin{eqnarray*}	
Q(\Theta|\Theta^{k})&=&E_{\Theta^{k}}[L(\Theta,P|Y)]\\
Q(L|M,N,Y)&=&\sum_{i=1}^n\sum_{j=1}^m\quad Q(L_{ij}|m_{ij}^{k},n_{ij}^{k},y_{ij})\\
&=&\sum_{i=1}^n\sum_{j=1}^m\quad E_{m_{ij}^{k},n_{ij}^{k}}[L_{ij}|y_{ij}]\\
&=&\sum_{i=1}^n\sum_{j=1}^m\quad Q_{ij}\\
\end{eqnarray*}

From the full probability function above, we can derive the posterior distribution of $P_{ij}$
\begin{eqnarray*}	
P(P_{ij}|m_{ij},n_{ij},y_{ij}) &\propto& P_{ij}^{y_{ij}+m_{ij}^k-1}(1-P_{ij})^{n_{ij}^k-y_{ij}}\\
&\sim &\mbox{Beta}\left(y_{ij}+m_{ij}^k,n_{ij}^k-y_{ij}+1\right) 
\end{eqnarray*}
Therefore, $Q$ can be expressed as:
\begin{eqnarray*}
Q(L|M,N,Y)&=&\sum_{i=1}^n\sum_{j=1}^m\quad Q_{ij}\\
&=& \sum_{i=1}^n\sum_{j=1}^m\quad E_{m_{ij}^{k},n_{ij}^{k}}[L_{ij}|y_{ij}]\\
&=& \sum_{i=1}^n\sum_{j=1}^m\quad(y_{ij}+m_{ij}-1)\left(\psi\left(y_{ij}+m_{ij}^{k}\right)
 -\psi\left(n_{ij}^{k}+m_{ij}^{k}+1\right)\right)\\
& & + (n_{ij}-y_{ij})\left(\psi\left(n_{ij}^{k}-y_{ij}+1\right)
 -\psi\left(n_{ij}^{k}+m_{ij}^{k}+1\right)\right)\\
&=& \sum_{i=1}^n\sum_{j=1}^m\quad(y_{ij}+m_{ij}-1)\psi\left(y_{ij}+m_{ij}^{k}\right)\\
& & + \sum_{i=1}^n\sum_{j=1}^m\quad(n_{ij}-y_{ij})\psi\left(n_{ij}^{k}-y_{ij}+1\right)\\
& & - \sum_{i=1}^n\sum_{j=1}^m\quad(m_{ij}+n_{ij}-1)\psi\left(n_{ij}^{k}+m_{ij}^{k}+1\right)\\
\mbox{Where } \psi(x)&=&\frac{\Gamma(x)^{'}}{\Gamma(x)}
  ~ \mbox{ is a digamma function}
\end{eqnarray*}

\noindent
{\bf M-step} \\
As we have the constraint
$\sum_{j=1}^m\beta_j=0$, we can express the Q function as follows:
\begin{enumerate}
\item
\begin{eqnarray*}
Q(L|M,N,Y)&=& \sum_{i=1}^n\sum_{j=1}^{m-1}m_{ij}SM_{ij}+\sum_{i=1}^n SM_{im}\left(\frac{\alpha_j\theta_i+\sum_{j=1}^{m-1}\beta_j}{2}\right)\\
& & + \sum_{i=1}^n\sum_{j=1}^{m-1}n_{ij}SN_{ij}+\sum_{i=1}^n SN_{im}\left(\frac{-\alpha_j\theta_i-\sum_{j=1}^{m-1}\beta_j}{2}\right)\\
\mbox{Where } SM_{ij}&=& \psi(y_{ij}+m_{ij}^{k})-\psi(n_{ij}^{k}+m_{ij}^{k}+1)\\
SN_{ij}&=& \psi(n_{ij}^{k}-y_{ij}+1)-\psi(n_{ij}^{k}+m_{ij}^{k}+1)
\end{eqnarray*}

The partial derivative of $Q$ with respect to $\theta_i,\beta_j$ is as follow:
\item
\begin{eqnarray*}
\frac{\partial Q}{\partial \theta_i}
 &=& \sum_{j=1}^{m}\frac{1}{2}m_{ij}SM_{ij}-\sum_{j=1}^{m}\frac{1}{2}n_{ij}SN_{ij}\\
\frac{\partial^2 Q}{\partial \theta_i^2}&=&
\sum_{j=1}^{m}\frac{1}{4}m_{ij}SM_{ij}+\sum_{j=1}^{m}\frac{1}{4}n_{ij}SN_{ij}\\
\frac{\partial Q}{\partial \beta_j}&=&
-\sum_{i=1}^{N}\frac{1}{2}m_{ij}SM_{ij}+\sum_{i=1}^{N}\frac{1}{2}m_{im}SM_{im}
+\sum_{i=1}^{N}\frac{1}{2}n_{ij}SN_{ij}-\sum_{i=1}^{N}\frac{1}{2}n_{im}SN_{im}\\
\frac{\partial^2 Q}{\partial \beta_j^2}&=&
\sum_{i=1}^{N}\frac{1}{4}m_{ij}SM_{ij}+\sum_{i=1}^{N}\frac{1}{4}m_{im}SM_{im}
+\sum_{i=1}^{N}\frac{1}{4}n_{ij}SN_{ij}+\sum_{i=1}^{N}\frac{1}{4}n_{im}SN_{im}
\end{eqnarray*}

Based on the partial derivatives, we can use the gradient ascent algorithm to update
the parameters $\Theta$ in each iteration.
\item
\begin{eqnarray*}
\theta_i^{k+1}&=&\theta_i^{k}+\alpha\frac{\partial Q}{\partial \theta_i}_{\Theta^{k}}\\
\beta_j^{k+1}&=&\beta_j^{k}+\alpha\frac{\partial Q}{\partial \beta_j}_{\Theta^{k}}\\
\end{eqnarray*}
\end{enumerate}

\section{Assigning weights to classifiers}

As the estimators of $\theta$ reflect classifiers' ability, we should assign classifier i a weight according to the estimation of $\theta_i$. The estimator, however, can be either positive or negative. Thus, we employ the following transformation to decide the value of weight:

\[w_i=\frac{\exp{(\theta_i)}}{\sum_{k=1}^n\exp{(\theta_k)}}\]
The weights clearly illustrate the order of classifiers' ability and are normalized by the above formula.

\section{Results}

\subsection{Results of the inference}

A simulation study is presented below to show the estimation of the parameters from two 10 by 1000
datasets, which are generated from different parameter settings. The difference lies in the
parameters' distributions.

First, we apply the three models to estimate the parameters of the two datasets generated from two
different sets of parameters. Both settings contain 10 samples with 1000 classifiers, which have fixed
difficulty and ability. In the first dataset, all parameters are sampled from a normal distribution so
that they meet the assumption of the first two models. However, the distributions of the parameters
in the second dataset vary. The difficulty is manually set to have a large variance and a relatively
large range, and the ability is sampled from a highly skewed gamma distribution that the second dataset
cannot satisfy the first two models' assumption. Thus, we can compare the accuracies of the two models
in different situations.

Four measures were computed to evaluate the performance of the three models. $\theta$ is the real parameter
and $\theta^{\mbox{est}}$ is an estimate of the parameter:

\begin{enumerate}
\item 
 \[ \mbox{Correlation Corr} = \frac{N\sum_{i=1}^{N}\theta_i^{\mbox{est}}\theta_i
-\sum_{i=1}^{N}\theta_i^{\mbox{est}}\sum_{i=1}^{N}\theta_i}{N\sigma_{\mbox{est}}\sigma} \]
\item
\[ \mbox{Mean Square Error MSE } = \frac{\sum_{i=1}^{N}(\theta_{i}^{\mbox{est}}-\theta_{i})^2}{N} \]
\item
\[ \mbox{Mean Absolute Error MAE } = \frac{\sum_{i=1}^{N}\|(\theta_{i}^{\mbox{est}}-\theta_{i})\|}{N}\]
\item
\[ \mbox{Variance ratio VR}  = \frac{\sigma_{\mbox{est}}^2}{\sigma^2} \]
\end{enumerate}

The correlation is to test the linear relationship between the parameter estimates and real parameters. The MSE and MAE measure the precision of the estimation and the Variance ratio illustrates the comparative stability. In Table \uppercase\expandafter{\romannumeral1} and Table \uppercase\expandafter{\romannumeral3}, we listed the previous four measures of the estimators (classifiers' abilities and problems' difficulties) obtained from different models.  In Talbe \uppercase\expandafter{\romannumeral2} and Table \uppercase\expandafter{\romannumeral4}, we show the real parameters and their estimated values from three different models. 

$\newline$
\begin{table}
\centering
\caption{ Measures for dataset 1}
\vspace{.1in}
\begin{tabular}{cccccc}
\hline
Meode l& Parameters& Correlation& MSE& MAE& VR\\
\hline
Model 1& $\theta$ (Classifiers' ability)& 0.86& 0.97& 0.96& 0.97\\
Model 2& $\theta$ (Classifiers' ability)& 0.88& 0.31& 0.37& 0.88\\
Model 3& $\theta$ (Classifiers' ability)& 0.87& 0.91& 0.96& 1.9\\
Model 1& $\beta$ (Samples' difficulty)& 0.97& 0.007& 0.16& 1.15\\
Model 2& $\beta$ (Samples' difficulty)& 0.99& 0.003& 0.05& 0.99\\
Model 3& $\beta$ (Samples' difficulty)& 0.99& 0.016& 0.12& 1.36\\
\hline
\end{tabular}
\end{table}

\begin{table}
\centering
\caption{Real and estimated parameters for dataset 1}
\vspace{.1in}
\begin{tabular}{ccccc}
\hline
Parameter&Parameters&Model 1&Model 2 &Model 3\\
Component&Value&Estimation&Estimation&Estimation\\
\hline
$\beta_1$& -1.024& -0.94& -0.953& -1.107 \\
$\beta_2$& -0.934& -0.99& -0.91& -1.08 \\
$\beta_3$& -0.694& -0.732& -0.709& -0.869 \\
$\beta_4$& -0.464& -0.56& -0.514& -0.579 \\
$\beta_5$& 0.356& 0.472& 0.408& 0.503 \\
$\beta_6$& 0.336& 0.336& 0.319& 0.382\\
$\beta_7$& 0.616& 0.694& 0.576& 0.722\\
$\beta_8$& 0.806& 0.85& 0.757& 0.956\\
$\beta_9$& -0.074& -0.171& -0.145& -0.172\\
$\beta_{10}$& 1.076& 1.23& 1.141& 1.243\\
\hline
\end{tabular}
\end{table}

\begin{center}
\begin{table}
\centering
\caption{ Measures for dataset 2}
\vspace{.1in}
\begin{tabular}{cccccc}
\hline
Meode l& Parameters& Correlation& MSE& MAE& VR\\
\hline
Model 1& $\theta$ (Classifiers' ability)& 0.63& 23& 4.2& 0.87\\
Model 2& $\theta$ (Classifiers' ability)& 0.87& 21.22& 3.9& 0.81\\
Model 3& $\theta$ (Classifiers' ability)& 0.93& 19.33& 3.1&0.95\\
Model 1& $\beta$ (Samples' difficulty)& 0.86& 15.75& 3.53& 0.8\\
Model 2& $\beta$ (Samples' difficulty)& 0.93& 12.64& 2.58& 0.18\\
Model 3& $\beta$ (Samples' difficulty)& 0.98& 1.716& 1.13& 1.08\\
\hline
\end{tabular}
\end{table}
\end{center}

\begin{center}
\begin{table}
\centering
\caption{Real and estimated parameters for dataset 2}
\vspace{.1in}
\begin{tabular}{ccccc}
\hline
Parameter&Parameters&Model 1&Model 2&Model3\\
Component&Value&Estimation&Estimation&Estimation\\
\hline
$\beta_1$& -9.075& -4.35& -3.256& -9.824 \\
$\beta_2$& -2.485& -1.3& -1.296& -2.68\\
$\beta_3$& -3.395& -1.2& -2.24& -3.85 \\
$\beta_4$& -1.015& -2.2& -0.192& -0.203 \\
$\beta_5$& -0.075& 0.461& 1.593& 1.248 \\
$\beta_6$& 0.165& 1.36& 0.96& 1.628\\
$\beta_7$& 4.035& 3.194& 3.045& 5.591\\
$\beta_8$& 6.485& 3.85& 3.783& 6.998\\
$\beta_9$& -6.395& -3.171& -2.904& -8.462\\
$\beta_{10}$& 11.755& 3.23& 3.534& 9.553\\
\hline
\end{tabular}
\end{table}
\end{center}

As we can see from the result, model 1 and model 2 perform well on the first dataset while model 3 beats
the others on the second dataset. The reason is because that first two models tend to give a relatively stable
solution due to the normal prior, which has the advantage to minimize the range between the estimators.
Thus, when the parameters are generated from a normal distribution, they fit well. In model 3, we didn't
have such an assumption, so the parameters can take any value of minimizing the loss function and the range
may be larger than that of the first two estimations. It clearly produces a better estimation when the
ranges of the real parameters are large. To compare the accuracy of the estimation, we calculate the error ratio for 
each parameter in both datasets. 
\[\mbox{Error Ratio}=\frac{(\theta^{\mbox{est}}-\theta)^2}{\mbox{Average MSE}}\]
\[\mbox{Average MSE}=\frac{\mbox{MSE}_1+\mbox{MSE}_2+\mbox{MSE}_3}{3}\]

\noindent
In Figure 2, we use a bar plot to compare the error ratio for all models on two datasets.

\begin{figure}[h]
	\centering  
	\includegraphics[width=0.66\linewidth]{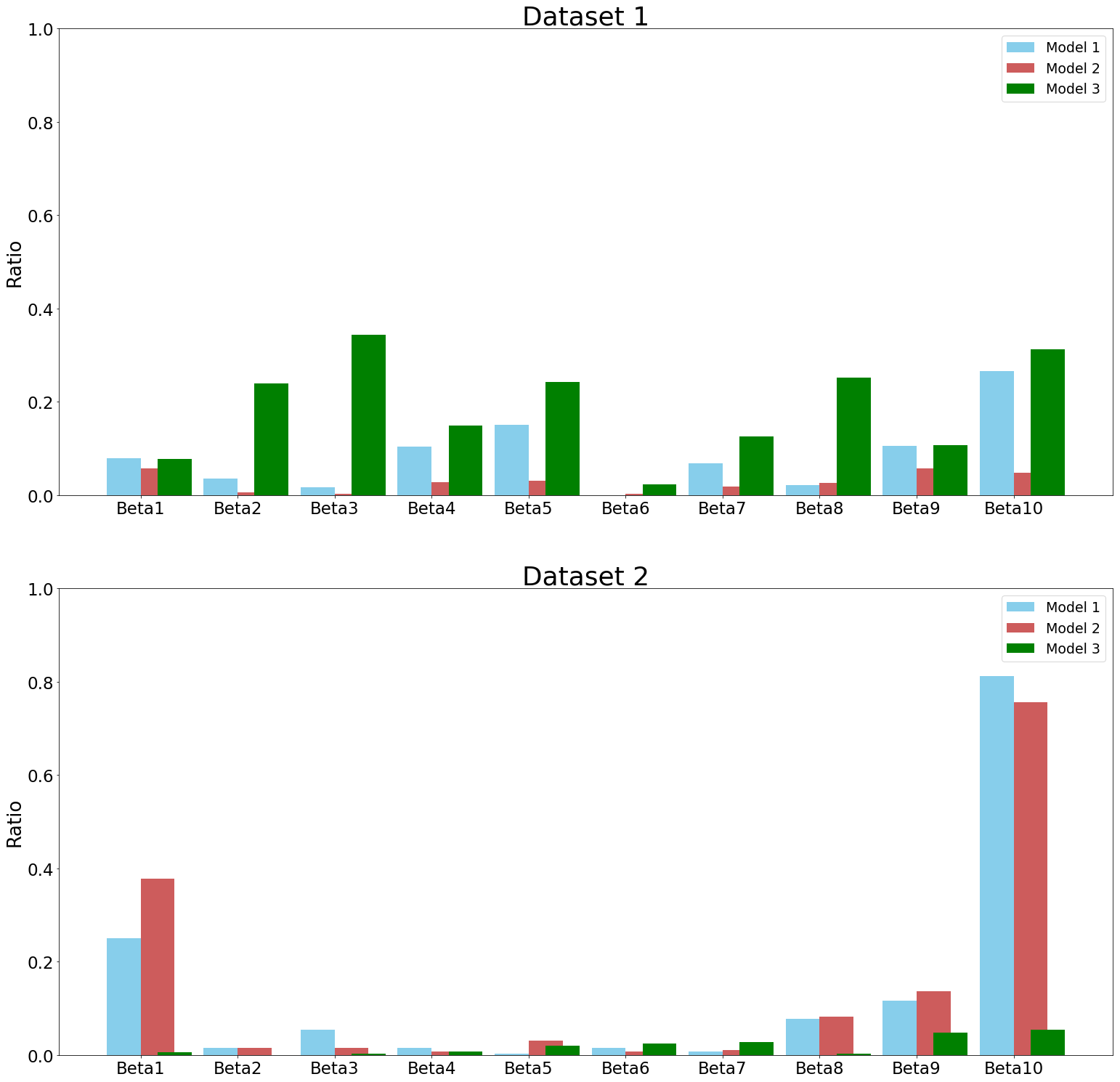}  
	\caption{The Error Ratio for the two datasets.}  
	\label{fig:mcmthesis-logo}   
\end{figure}

From the results, we can make a conclusion that real parameters and parameter estimates are highly
correlated. For both datasets, the correlations are always higher than 0.85, implying that the
parameter estimates hold a strong linear relationship with the true parameter values. Thus, it makes more
sense to keep the weights of different classifiers as ordinal consistent with the classifiers' abilities\cite{fang2014prioritizing}.

\subsection{Analysis of difficulty parameters}

As we mentioned before, an IRT ensemble can evaluate the classifiers' ability and samples' difficulty simultaneously. To better understand the model, we first show what the parameters can reflect, and how they can affect the classification. Then we will illustrate the performance of our method on different datasets.

We depict the sample points from the chess board dataset to show how the IRT ensemble model evaluates samples' difficulty. Figure 3 and Figure 4 illustrate the dual character of location and difficulty, which is shown by the size of the sample points. The larger the point, the larger the difficulty parameter. Points in the same block share the same color in the original chess board. We first show how the estimated values change with increased iterations. In Figure 3, we constructed 500 base classifiers. As we increase the number of iterations, difference gradually increases. The outcome of the 100 iterations is similar, while the outcome of the 500 iterations is various, meaning each sample is well distinguished by its difficulty evaluated by a bunch of classifiers. Zooming in to the figure, we can find some rules. The IRT Ensemble method tends to assign a higher difficulty to the points that are closer to the boundary, and these points support the decision boundary in return. It is obvious that when the points are close to their counterparts with a distinct label, they are more likely to be misclassified. Only those classifiers which are powerful enough can correctly complete the task of difficult classification. Thus it makes sense to take their capability for constructing the classification boundary.

\begin{figure}[h]
	\centering  
	\includegraphics[width=0.66\linewidth]{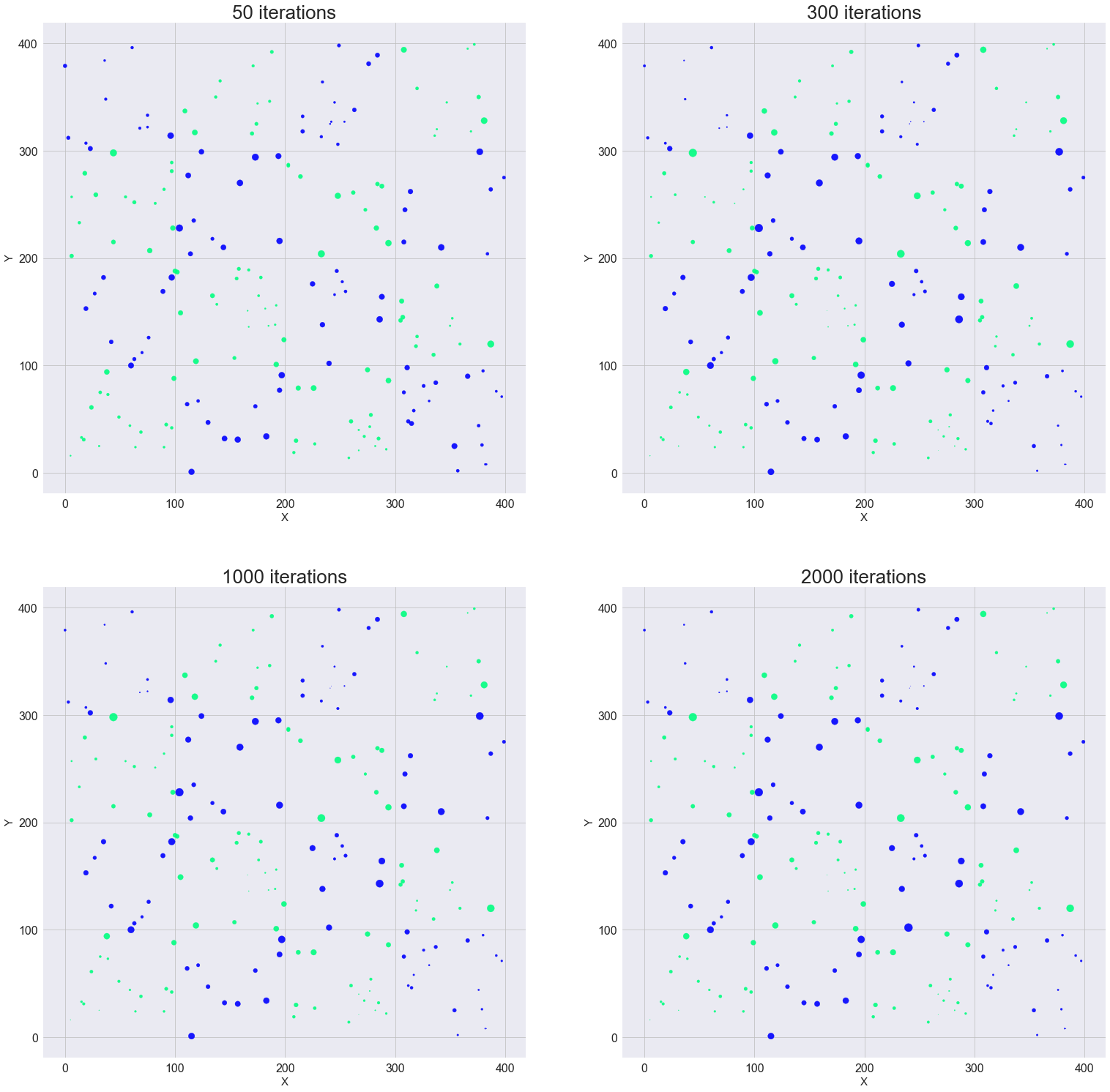}  
	\caption{500 Classifiers with different iterations. X and Y axes illustrate the position of the points, and different class labels are shown by distinct colors.}  
	\label{fig:mcmthesis-logo}   
\end{figure}

For some other ensemble methods, it proves that increasing the number of classifiers can improve the performance, which also applies to the IRT ensemble method. In Figure 4, we fixed the number of iterations to be 2000 and changed the number of classifiers. According to the experiment, when we increase the number of classifiers, those sample points constructing the boundary within the block will stand out while those inside the boundary will shrink to a dot. In order to distinguish the important sample points from others, more classifiers should be included to make a joint decision. The interesting bit comes when we increase the number of classifiers to a large enough value. In the last two subplots, the sizes of the points seem unchangeable. In many cases, increasing the decision size cannot guarantee improved performance. When we have sufficient classifiers, it will come across the bottleneck.

\begin{figure}[h]
	\centering  
	\includegraphics[width=0.66\linewidth]{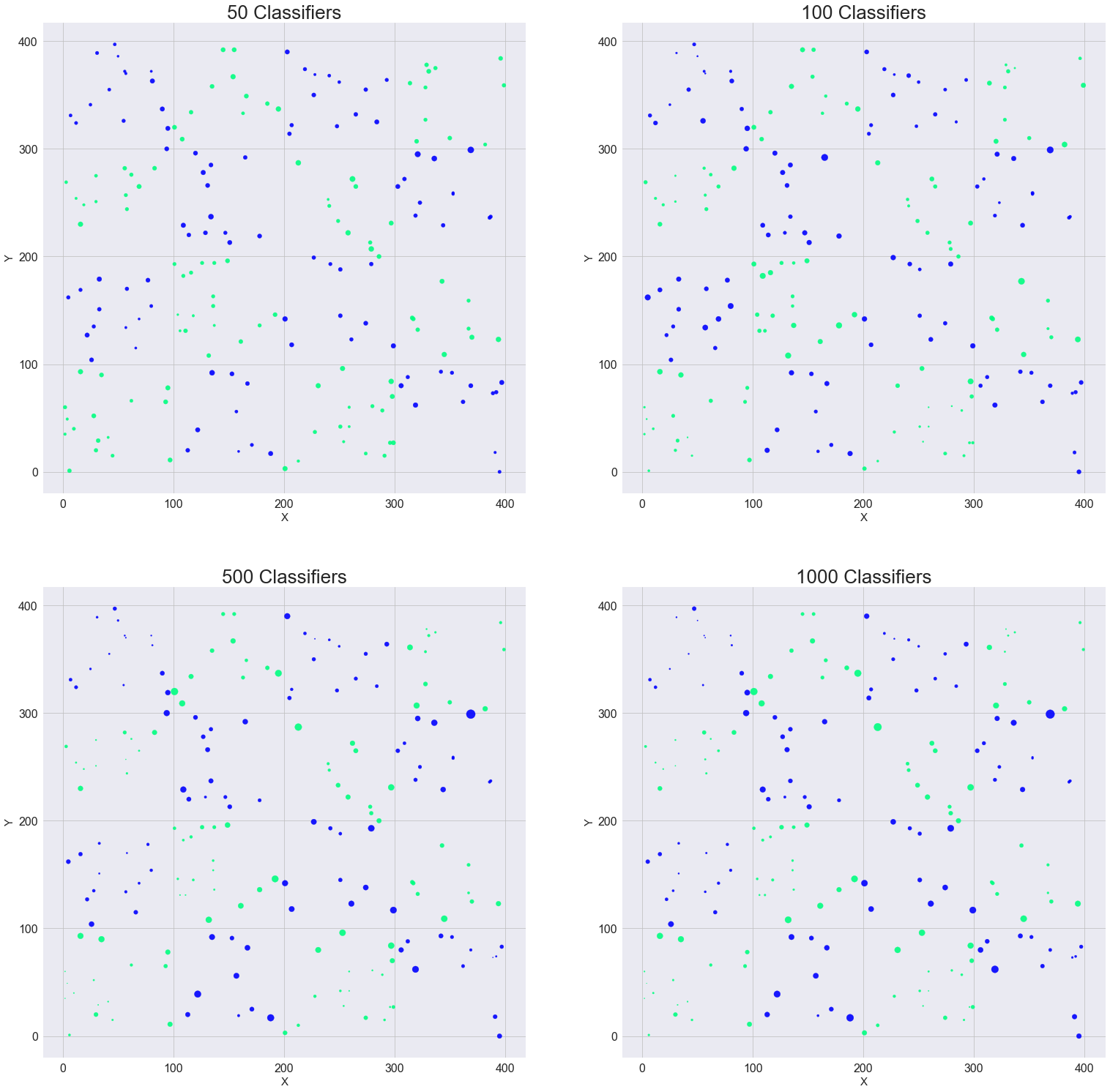}  
	\caption{Different number of classifiers. X and Y axes illustrate the position of the points, and different class labels are shown by distinct colors.}  
	\label{fig:mcmthesis-logo}   
\end{figure}

\subsection{Analysis of classification}

We collected 15 real datasets and 2 artificial datasets to compare our Model 2 with a single tree\cite{safavian1991survey}, random forest\cite{liaw2002classification}, bagging\cite{breiman1996bagging}, gradient boosting\cite{friedman2002stochastic}, LDA\cite{mika1999fisher}, and SVM\cite{suykens1999least}. We didn't make experiments for Model 1 because it is time consuming.  For all the ensemble methods, we used a single classification tree as the base classifier. When it comes to a single tree, the pruning option is necessary for preventing overfitting. However, we didn't implement the pruning algorithm for the base classifiers in bagging, gradient boosting and random forest because pruning decreases the variance but increases the bias. For all the ensemble algorithms, bootstrap can be used to construct various base tree structures, which can reduce the variance effectively. Thus in all the ensemble models, we used unpruned trees as the base classifiers to account for bias and then used bootstrap to reduce the variance.

Most of the datasets summarized in Table \uppercase\expandafter{\romannumeral5} are from the UCI datasets. As our model is constructed using the base classifiers, it is suitable for all kinds of features as long as the base model is adequate for the data. In order to show a generalization of our method, we intentionally selected some datasets that containing both the categorical features as well as the continuous features. For all the continuous features, normalization is performed. One-hot encoding is also a must for all the nominal features. We conducted all the experiments on Python 3.6 platform. To compare the accuracy of these methods, we randomly split the dataset to 10 fold and set the test set proportional to 0.3. A simulation of each setting was performed 30 times for each dataset. In order to compare the accuracies of various methods, we set the number of trees in each ensemble algorithm to 500.
\begin{table}
\centering
\caption{Dataset information}
\vspace{.1in}
\begin{tabular}{cccccc}
\hline

Dataset & Observations & Continuous & Discrete& Class & Source\\
&&Features&Features&&\\
\hline
IRIS & 150 & 4 & 0 & 3 & UCI\\ 
Bld & 345 & 6 & 0 & 2 & UCI\\ 
Spe & 267 & 44 & 0 & 2 & UCI\\ 
Glass & 214 & 9 & 0 & 6 & UCI\\ 
Veh & 846 & 18 & 0 & 4 & UCI\\ 
Checkboard & 160000 & 2 & 0 & 2 & Artificial\\
BTD & 106 & 9 & 0 & 4 & UCI\\ 
IPLD & 583 & 8 & 1 & 2 & UCI\\ 
Haberman & 306 & 3 & 0 & 2 & UCI\\ 
Ionos & 351 & 32 & 2 & 2 & UCI\\ 
Multiangle & 160000 & 2 & 0 & 2 & Tensorflow\\ 
Balance & 625 & 0 & 4 & 2 & UCI\\ 
AUS & 690 & 5 & 9 & 2 & UCI\\ 
ECOLI & 336 & 7 & 0 & 6 & UCI\\ 
LEN & 24 & 0 & 3 & 3 & UCI\\ 
TAE & 151 & 0 & 5 & 3 & UCI\\ 
LC & 33 & 0 & 56 & 3 & UCI\\ 
LSVT & 126 & 310 & 0 & 2 & UCI\\
SCADI & 70 & 205 & 0 & 8 & UCI\\\hline

\end{tabular}
\end{table}

The result of the average accuracy is in Table \uppercase\expandafter{\romannumeral6}. We highlighted the best two results and the worst result for each data set. From the accuracy table, it seems that Model 2, random forest and gradient boosting perform well in general. However, for some kinds of data sets, gradient boosting fails to recognize that pattern and yield the worst result\cite{bauer1999empirical}. The weakness of gradient boosting is reported in some papers before . The performance of SVM also greatly fluctuated\cite{li2015prediction}.

\begin{table}
\centering
\caption{Average of accuracy}
\vspace{.1in}
\begin{tabular}{cccccccc}
\hline

DataSet & Model 2 & RF & GBDT & SVM & Tree & Bagging & LDA\\ \hline
IRIS & \bf{0.967} & \bf{0.962} & 0.945 & 0.932 & 0.947 & 0.954 & \bf{0.88}\\ 
Bld & \bf{0.703} & 0.69 & \bf{0.693} & 0.68 & 0.62 & 0.66 & \bf{0.59}\\ 
Spe & \bf{0.91} & \bf{0.91} & 0.88 & \bf{0.828} & 0.861 & \bf{0.911} & 0.865\\ 
Glass & 0.729 & \bf{0.766} & \bf{0.77} & 0.63 & 0.72 & 0.72 & \bf{0.59}\\ 
Veh & 0.74 & 0.73 & \bf{0.75} & \bf{0.76} & 0.63 & 0.68 & \bf{0.53}\\ 
Checkboard &\bf{0.88} & 0.82 & \bf{0.96} & \bf{0.5} & 0.58 & 0.87 & \bf{0.5}\\ 
BTD & 0.85 & 0.875 & \bf{0.88} & \bf{0.89} & \bf{0.82} & 0.86 & 0.85\\ 
IPLD & \bf{0.723} & \bf{0.712} & 0.7 & 0.71 & \bf{0.68} & 0.7 & \bf{0.68}\\ 
Harman & 0.677 & 0.668 & \bf{0.65} & \bf{0.74} & 0.71 & \bf{0.73} & \bf{0.73}\\ 
Ionos & \bf{0.928} & \bf{0.926} & 0.92 & 0.886 & 0.888 & 0.923 & \bf{0.855}\\ 
Multiangle & \bf{0.92} & 0.9 & \bf{0.96} & 0.51 & 0.83 & 0.83 & \bf{0.49}\\ 
Balance & \bf{0.815} & \bf{0.849} & 0.623 & 0.78 & \bf{0.59} & 0.65 & 0.73\\ 
Aus & \bf{0.858} & \bf{0.865} & 0.81 & 0.76 & 0.82 & 0.83 & \bf{0.68}\\ 
ECOLI & \bf{0.833} & 0.755 & \bf{0.65} & \bf{0.87} & 0.8 & 0.81 & 0.829\\ 
Len & \bf{0.738} & \bf{0.706} & \bf{0.725} & 0.717 & 0.708 & 0.72 & \bf{0.725}\\ 
TAE &\bf{0.565} & 0.468 & \bf{0.582} & 0.511 & \bf{0.37} & 0.43 & 0.44\\ 
LC & \bf{0.5} & \bf{0.484} & \bf{0.5} & 0.45 & 0.43 & 0.43 & \bf{0.2}\\ \hline

\end{tabular}
\end{table}

Bagging generally performs better than the tree model. Although bagging is not the best, it is more stable than gradient boosting in some cases. It is noted that Model 2 is within the scope of a weighted voting model, which extends from the bagging strategy. Thus, we can explain the reason why Model 2 is more stable than the gradient boosting method.

A win table\cite{chen2014canonical} summarizes the comparison in Table \uppercase\expandafter{\romannumeral7}. In the win table, $a_{i,j}$ illustrates the frequency that method $j$ gives a higher accuracy than method $i$. For instance, $a_{1,3}$ equals 11 means in total 17 comparisons between Model 2 and gradient boosting, Model 2 produces more accurate or the same result than gradient boosting in 11
datasets. This table shows every pairwise comparison in detail. In order to rank the methods, we need to calculate the goal difference from the win table, subtracting the frequency of loss from the frequency of wins for each model. The frequency of wins can be obtained by summing within the row, while the frequency of losses can be obtained within the column for each model. From there, the goal difference can be calculated. The result is shown in Table \uppercase\expandafter{\romannumeral8}. It is clear that Model 2 has an overwhelming superiority over others.

\begin{table}
\centering
\caption{Win Table}
\vspace{.1in}
\begin{tabular}{cccc}
\hline

Models & WinLoss & Win & Loss\\ \hline
Model 2 & 53  & 96  & 43 \\ 
RandomForest & 23  & 80  & 57 \\ 
GBDT & 19  & 79  & 60 \\ 
SVM & 5  & 71  & 66 \\ 
Tree & -41  & 50  & 91 \\ 
Bagging & -5  & 68  & 73 \\ 
LDA & -54  & 44  & 98\\ \hline

\end{tabular}
\end{table}

We also conducted an experiment to investigate how the ensemble size affects the prediction with 13 datasets
(4 datasets were discarded because of the failure of computation when the ensemble size is too small). We still used the same method for getting the accuracy and 20 repetitions were conducted for each sample size, which are averaged to calculate the t~statistics. Figure 5 we shows the boxplot. When comparing our method with random forest, gradient boosting and SVM, we conclude that the gain of our model tends to be enhanced as the ensemble size increases. It seems that our model has a good potential to improve the accuracy if the ensemble size is large enough.

\begin{figure}[h]
	\centering  
	\includegraphics[width=0.66\linewidth]{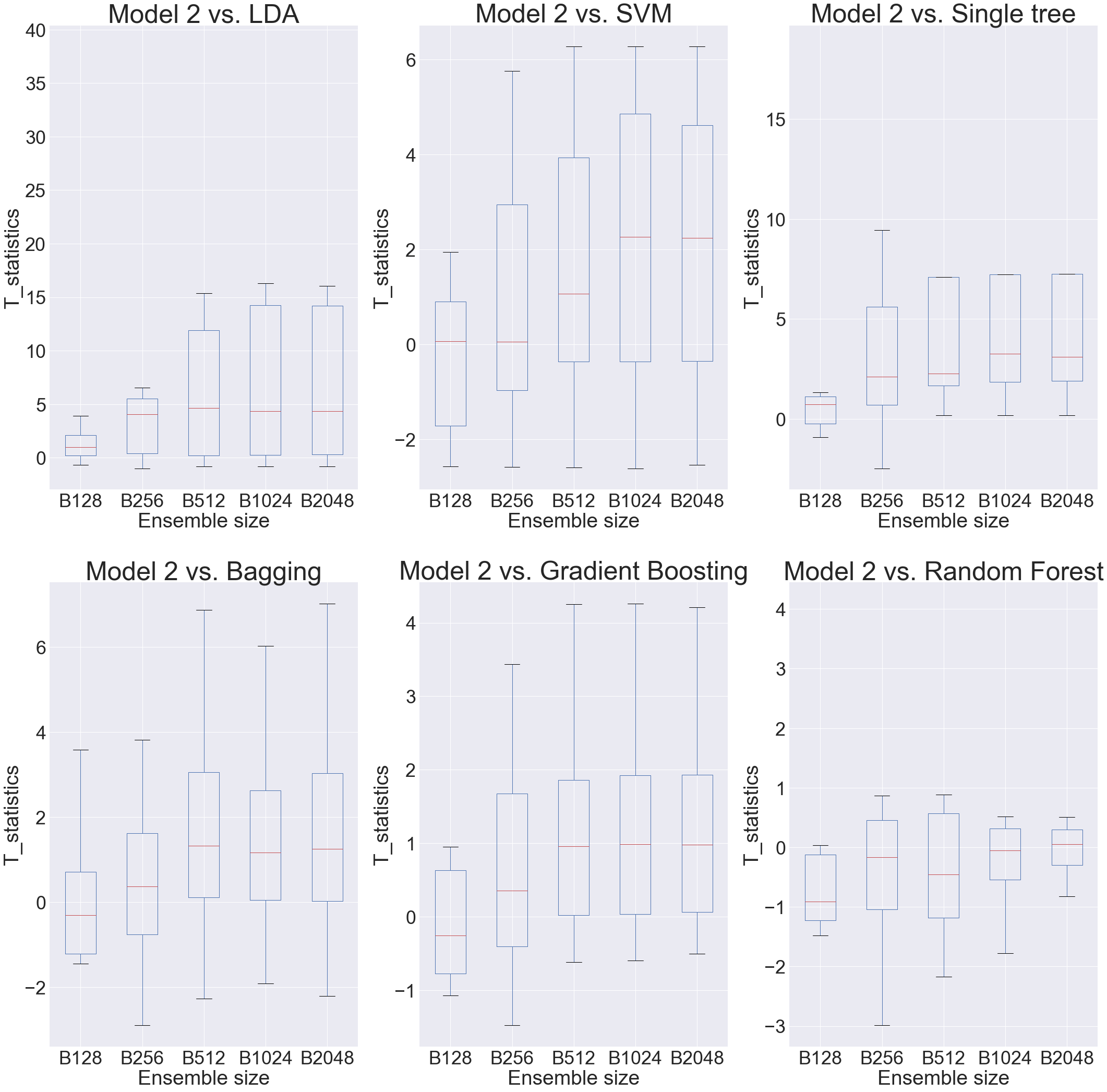}  
	\caption{Comparison with LDA,SVM,Single tree,Bagging,Gradient and Random Forest}  
	\label{fig:mcmthesis-logo}   
\end{figure}

Model 3 has predominant advantage when applying to some datasets. We illustrate the cumulative accuracy on 4 cases in Figure 6. In theses datasets, samples contain many attributes compare to the sample size, which means they include a lot of redundant information. Consequently, the samples' difficulties vary and there may exist a small subsample contributing a lot for constructing the decision boundary. It is hard for most of the classifiers to detect the important variables. Only a small subset of classifiers are powerful. Thus, the distribution of the classifiers are no longer symmetric and the variance of the classifiers' abilities increases. Model 3 is more powerful than Model 2 when the variance of the parameters is large enough, and higher weights are assigned for the strong classifiers. Thus, it can outperform other methods in these cases. We found that Model 3 can consistently produce a higher accuracy than random forest and gradient boosting. However, when the difficulty of each sample is similar, Model 2 tends to perform better.

\begin{figure}[h]
	\centering  
	\includegraphics[width=0.66\linewidth]{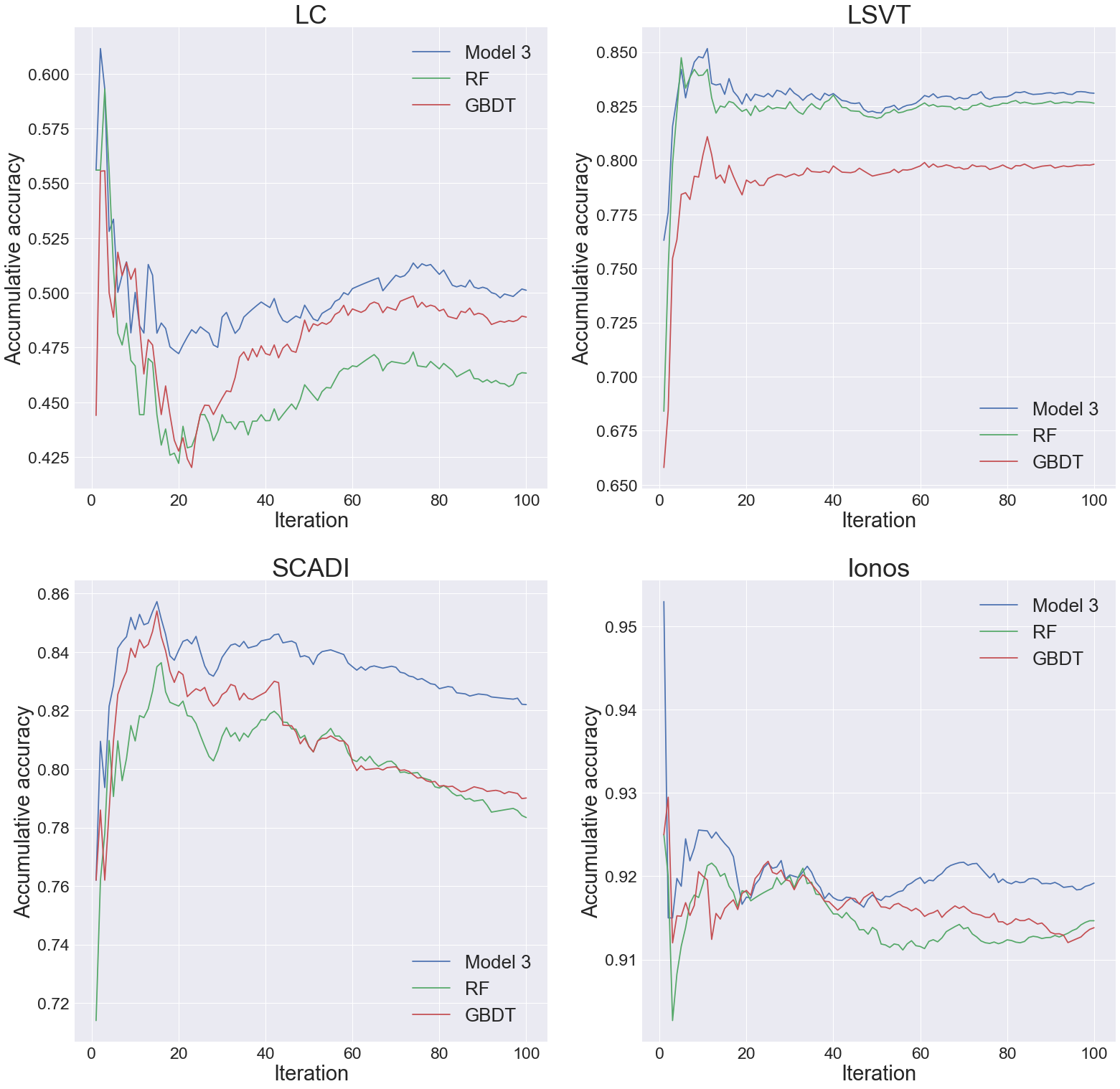}  
	\caption{Cumulative accuracy for Model 3 on 4 different datasets}  
	\label{fig:mcmthesis-logo}   
\end{figure}

\section{Conclusion}

In this paper, we proposed the IRT ensemble, a weighted majority voting method focusing on the classifiers that can correctly deal with the hard-to-classify problems, by adopting the item response theory. The classifying boundaries are constructed by the points that are frequently misclassified and higher weights are assigned to the classifiers with higher abilities. We also proposed three models to estimate the ability parameters and introduced the assumptions behind the models.

For the performance of the models, we analyzed them in two stages. First, we evaluated their accuracy in the estimation of parameters. We concluded that Model 1 and Model 2 perform well when the variance of the parameters are small, while Model 3 is more suitable when the parameters vary. We also explained how the lengths of the Markov chains and the number of classifiers would affect the estimation of samples' difficulty. The chessboard dataset also provides us an intuitive explanation about the idea behind the IRT ensemble algorithm. Finally, we implemented an experiment with Model 2 using 19 datasets and compared the performance with other classification methods. We showed that the advantage of model 2 is enhanced with the increased ensemble size compared to LDA, SVM, single tree, bagging, and gradient boosting. It showed compatible performance with random forest. Finally, we found the Model 3 has an edge in high dimensional datasets.

Future work includes combining the Model 3 with kernel methods. Another modification is to introduce the Beta model, which is widely used in the network analysis.

\bibliographystyle{plain}
\bibliography{ci}

\end{document}